%% file: main.tex
\documentclass[11pt,a4paper,dvipsnames]{article}
\usepackage[hyperref]{acl2020}
\usepackage{times}
\usepackage{latexsym}
\usepackage{graphicx}
\usepackage{xspace}
\usepackage{multirow, makecell}
\usepackage{amsmath}
\usepackage{amsthm,bm}
\usepackage{microtype}
\usepackage{subfiles}
\usepackage{subcaption}
\usepackage{lipsum}
\usepackage{multirow}
\usepackage{arydshln}
\usepackage{booktabs}
\usepackage{amsfonts}
 \usepackage{todonotes}
\usepackage{enumitem}
\usepackage{amssymb}
\usepackage{graphicx}
\usepackage{wasysym}
\usepackage{makecell}

\newcommand{\tanda}{\textsc{TandA}\xspace}
\newcommand{\astwo}{$\text{AS2}$\xspace}

\newcommand{\ra}[1]{\renewcommand{\arraystretch}{#1}}
\newcommand{\STAB}[1]{\begin{tabular}{@{}c@{}}#1\end{tabular}}
\newcommand{\Alexa}{Amazon Alexa\xspace}

\definecolor{ylw}{HTML}{B29334}
\definecolor{ble}{HTML}{587BB0}
\definecolor{rdd}{HTML}{923936}
\definecolor{prp}{HTML}{755386}

\aclfinalcopy %
\def\aclpaperid{2317} %

\title{The Cascade Transformer: an Application for\\ Efficient Answer Sentence Selection}

\author{Luca Soldaini \\
  Amazon Alexa  \\ Manhattan Beach, CA, USA \\
  \texttt{lssoldai@amazon.com} \\\And
  Alessandro Moschitti \\
  Amazon Alexa  \\ Manhattan Beach, CA, USA \\
  \texttt{amosch@amazon.com} \\}

\date{}

\begin{document}
\maketitle

\subfile{sections/abstract}
\subfile{sections/intro}

\subfile{sections/related-works}

\subfile{sections/task-defitnion}

\subfile{sections/methodology}

\subfile{sections/experimental-setup}

\subfile{sections/results}

\subfile{sections/conclusions}

\bibliography{main,alt,aaai2020}
\bibliographystyle{acl_natbib}

\end{document}

%% file: sections/abstract.tex
\begin{abstract}
Large transformer-based language models have been shown to be very effective in many classification tasks.
However, their computational complexity prevents their use in applications requiring the classification of a large set of candidates.
While previous works have investigated approaches to reduce model size, relatively little attention has been paid to techniques to improve batch throughput during inference.
In this paper, we introduce the Cascade Transformer, a simple yet effective technique to adapt transformer-based models into a cascade of rankers.
Each ranker is used to prune a subset of candidates in a batch, thus dramatically increasing throughput at inference time.
Partial encodings from the transformer model are shared among rerankers, providing further speed-up.
When compared to a state-of-the-art transformer model, our approach reduces computation by 37\% with almost no impact on accuracy, as measured on two English Question Answering datasets.
\end{abstract}

%% file: sections/intro.tex
\section{Introduction}
\label{sec:introduction}

Recent research has shown that transformer-based neural networks can greatly advance the state of the art over many natural language processing tasks.
Efforts such as BERT \cite{devlin2019bert}, RoBERTa \cite{liu2019roberta}, XLNet \cite{DBLP:journals/corr/abs-1901-02860}, and others have led to major advancements in several NLP subfields.
These models are able to approximate syntactic and semantic relations between words and their compounds by pre-training on copious amounts of unlabeled data \cite{clark2019bertattention,jawahar2019structure}.
Then, they can easily be applied to different tasks by just fine-tuning them on training data from the target domain/task \cite{liu2019linguistic,peters2019tune}.
The impressive effectiveness of transformer-based neural networks can be partially attributed to their large number of parameters (ranging from 110 million for ``base'' models to over 8 billion \cite{shoeybi2019megatronlm}); however, this also makes them rather expensive in terms of computation time and resources.
Being aware of this problem, the research community has been developing techniques to prune unnecessary network parameters \cite{lan2019albert,sanh2019distilbert} or optimize the transformer architecture \cite{zhang2018accelerating,xiao2019sharing}.

In this paper, we propose a completely different approach for increasing the efficiency of transformer models, which is orthogonal to previous work, and thus can be applied in addition to any of the methods described above.
Its main idea is that a large class of NLP problems requires choosing one correct candidate among many.
For some applications, this often entails running the model over hundreds or thousands of instances.
However, it is well-known that, in many cases, some candidates can be more easily excluded from the optimal solution \cite{Landi60}, i.e., they may require less computation.
In the case of hierarchical transformer models, this property can be exploited by using a subset of model layers to score a significant portion of candidates, i.e., those that can be \emph{more easily} excluded from search.
Additionally, the hierarchical structure of transformer models intuitively enables the re-use of the computation of lower blocks to feed the upper blocks.

Following the intuition above, this work aims at studying how transformer models can be cascaded to efficiently find the max scoring elements among a large set of candidates.
More specifically, the contributions of this paper are:

First, we build a sequence of rerankers $SR_N=\{R_1, R_2,...,R_N\}$ of different complexity, which process the candidates in a pipeline. Each reranker at position $i$ takes the set of candidates selected by $(i-1)$-th reranker and provides top $k_i$ candidates to the reranker of position $i+1$.
By requiring that $k_{i} < k_{i - 1} \ \ \forall i=1,\ldots, N-1$, this approach allows us to save computation time from the more expensive rerankers by progressively reducing the number of candidates at each step.
We build $R_i$ using transformer networks of 4, 6, 8, 10, and 12 blocks from RoBERTa pre-trained models.

Second, we introduce a further optimization on $SR_N$ to increase its efficiency based on the observation that
models $R_i$ in $SR_N$ process their input independently.
In contrast, we propose the Cascade Transformer (CT), a sequence of rerankers built on top of a single transformer model.
Rerankers $R_1, \ldots, R_N$ are obtained by adding small feed-forward classification networks at different transformer block positions;
therefore, the partial encodings of the transformer blocks are used as both input to reranker $R_i$, as well as to subsequent transformer encoding blocks.
This allows us to efficiently re-use partial results consumed by $R_i$ for rankers $R_{i+1}, \ldots, R_N$.

To enable this approach, the parameters of all rerankers must be compatible.
Thus, we trained CT in a multi-task learning fashion, alternating the optimization for different $i$, i.e., the layers of $R_i$ are affected by the back-propagation of its loss as well as by the loss of $R_{j}$, with $j\leq i$.

Finally, as a test case  for CT, we target Answer Sentence Selection (\astwo), a well-known task in the domain of Question Answering (QA).
Given a question and a set of sentence candidates (e.g., retrieved by a search engine), this task consists in selecting sentences that correctly answer the question.
We tested our approach on two different datasets: (\textit{i}) ASNQ, recently made available by \citet{garg2019tanda}; and (\textit{ii}) a benchmark dataset built from a set of anonymized questions asked to \Alexa.
Our code, ASNQ split, and models trained on ASNQ are publicly available.\footnote{\url{https://github.com/alexa/wqa-cascade-transformers}}

Our experimental results show that:
(\textit{i}) The selection of different $k_i$ for $SR_N$ determines different trade-off points between efficiency and accuracy. For example, it is possible to reduce the overall computation by 10\% with just 1.9\% decrease in accuracy.
(\textit{ii}) Most importantly, the CT approach largely improves over SR, reducing the cost by 37\% with almost no loss in accuracy. (\textit{iii}) The rerankers trained through our cascade approach achieve equivalent or better performance than transformer models trained independently.
Finally, (\textit{iv}) our results suggest that CT can be used with other NLP tasks that require candidate ranking, e.g., parsing, summarization, and many other structured prediction tasks.

%% file: sections/related-works.tex
\section{Related Work}
\label{sec:related-works}
In this section, we first summarize related work for sequential reranking of passages and documents, then we focus on the latest methods for AS2, and finally, we discuss the latest techniques for reducing transformer complexity.

\paragraph{Reranking in QA and IR}
The approach introduced in this paper is inspired by our previous work \cite{yoshi_et_al_2020}; there, we used a fast AS2 neural model to select a subset of instances to be input to a transformer model. This reduced the computation time of the latter up to four times, preserving most accuracy.

Before our paper, the main work on sequential rankers originated from document retrieval research.
For example, \citet{Wang-2011} formulated and developed a cascade ranking model that improved both top-k ranked effectiveness and retrieval efficiency.
\citet{Dang:2013:TLR:2458181.2458228} proposed two stage approaches using a limited set of textual features and a final model trained using a larger set of query- and document-dependent features.
\citet{Wang:2016:FFC:2911451.2911515} focused on quickly identifying a set of good candidate documents that should be passed to the second and further cascades.
\citet{Gallagher:2019:JOC:3289600.3290986} presented a new general framework for learning an end-to-end cascade of rankers using back-propagation.
\citet{Asadi:2013:ETC:2484028.2484132} studied effectiveness/efficiency trade-offs with three candidate generation approaches.
While these methods are aligned with our approach, they  target document retrieval, which is a very different setting.
Further, they only used linear models or simple neural models.
\citet{Agarwal:2012:LRR:2396761.2396867} focused on \astwo, but just applied linear models.

\paragraph{Answer Sentence Selection (\astwo)}
In the last few years, several approaches have been proposed for \astwo.
For example, \citet{severyn2015learning} applied CNN to create question and answer representations, while others proposed inter-weighted alignment networks \citep{shen-etal-2017-inter,tran-etal-2018-context,DBLP:journals/corr/abs-1806-00778}.
The use of compare and aggregate architectures has also been extensively evaluated \citep{wang2016compare,Bian:2017:CMD:3132847.3133089,DBLP:journals/corr/abs-1905-12897}. This family of approaches uses a shallow attention mechanism over the question and answer sentence embeddings.
Finally, \citet{tayyar-madabushi-etal-2018-integrating} exploited fine-grained question classification to further improve answer selection.

Transformer models have been fine-tuned on several tasks that are closely related to \astwo.
For example, they were used for machine reading \cite{devlin2019bert,yang-etal-2019-end-end,Wang2019ToTO}, ad-hoc document retrieval \cite{DBLP:journals/corr/abs-1903-10972,macavaney2019cedr}, and semantic understanding \cite{Liu2019MultiTaskDN} tasks to obtain significant improvement over previous neural methods.
Recently, \citet{garg2019tanda} applied transformer models, obtaining an impressive boost of the state of the art for \astwo tasks.

\paragraph{Reducing Transformer Complexity}
The high computational cost of transformer models prevents their use in many real-word applications.
Some proposed solutions rely on leveraging knowledge distillation in the pre-training step, e.g., \cite{sanh2019distilbert}, or used parameter reduction techniques \cite{lan2019albert} to reduce inference cost.
However, the effectiveness of these approaches varies depending on the target task they have been applied to. %
Others have investigated methods to reduce inference latency by modifying how self-attention operates, either during encoding \cite{child2019generating,guo2019star}, or decoding \citep{xiao2019sharing,zhang2018accelerating}.
Overall, all these solutions are mostly orthogonal to our approach, as they change the architecture of transformer cells rather than efficiently re-using intermediate results.

With respect to the model architecture, our approach is similar to probing models\footnote{Also known as auxiliary or diagnostic classifiers.} \cite{adi2017finegrained,liu2019linguistic,hupkes2018visualisation,belinkov2017evaluating}, as we train classification layers based on partial encoding on the input sequence.
However, (\textit{i}) our intermediate classifiers are integral part of the model, rather than being trained on frozen partial encodings, and (\textit{ii}) we use these classifiers not to inspect model properties, but rather to improve inference throughput.

Our apporach also shares some similarities with student-teacher (ST) approaches for self-training \cite{yarowsky1995unsupervised,mcclosky2006reranking}.
Under this setting, a model is used both as a ``teacher'' (which makes predictions on unlabeled data to obtain automatic labels) and as a ``student'' (which learns both from gold standard and automatic labels).
In recent years, many variants of ST have been proposed, including treating teacher predictions as soft labels
\cite{hinton2015distilling}, masking part of the label \cite{clark2018semi}, or use multiple modules for the teacher \cite{zhou2005tri, ruder2018strong}.
Unlike classic ST approaches, we do not aim at improving the teacher models or creating efficient students; instead, we trained models to be used as sequential ranking components.
This may be seen as a generalization of the ST approach, where the student needs to learn a simpler task than the teacher.
However, our approach is significantly different from the traditional ST setting, which our preliminary investigation showed to be not very effective.

%% file: sections/task-defitnion.tex
\section{Preliminaries and Task Definition}
\label{sec:taskdef}

We first formalize the problem of selecting the most likely element in a set as a reranking problem; then, we define sequential reranking (SR); finally, we contextualize \astwo task in such framework.

\subsection{Max Element Selection}
\label{AS}

In general, a large class of NLP (and other) problems can be formulated as a max element selection task: given a query $q$ and a set of candidates $A=\{a_1,..,a_n\}$, select $a_j$ that is an optimal element for $q$.
We can model the task as a selector function $\pi: Q \times \mathcal{P}(A) \rightarrow A$, defined as $\pi(q,A) = a_j$, where $\mathcal{P}(A)$ is the powerset of $A$, $j={\tt argmax}_i \hspace{.3em} p(q,a_i)$, and $p(q,a_i)$ is the probability of $a_i$ to be the required element.
$p(q,a_i)$ can be estimated using a neural network model.
In the case of transformers, said model can be optimized using a point-wise loss, i.e., we only use the target candidate to generate the selection probability.
Pair-wise or list-wise approaches can still be used \cite{Bian:2017:CMD:3132847.3133089}, but (\textit{i}) they would not change the findings of our study, and (\textit{ii}) point-wise methods have been shown to achieve competitive performance in the case of transformer models.

\subsection{Search with Sequential Reranking (SR)}
\label{SR}

Assuming that no heuristics are available to pre-select a subset of most-likely candidates, max element selection requires evaluating each sample using a relevance estimator.
Instead of a single estimator, it is often more efficient to use a sequence of rerankers to progressively reduce the number of candidates.

We define a reranker as a function $R: Q \times \mathcal{P}(A) \rightarrow \mathcal{P}(A)$, which takes a subset $\Sigma \subseteq A$, and returns a set of elements, $R(q,\Sigma) = \{a_{i1},...,a_{ik}\} \subset \Sigma$ of size $k$, with the highest probability to be relevant to the query. That is, $p(q,a)>p(q,b) \quad \forall a \in \Sigma,  \quad \forall b \in A-\Sigma$.

Given a sequence of rerankers sorted in terms of computational efficiency, $(R_1$,$R_2$, \dots ,$R_N)$, we assume that the ranking accuracy, $\mathcal{A}$ (e.g., in terms of MAP and MRR), increases in reverse order of the efficiency, i.e., $\mathcal{A}(R_j) > \mathcal{A}(R_i)$ iff $j>i$.
Then, we define a Sequential Reranker of order $N$ as the composition of $N$ rerankers: $SR_N(A) =R_N \circ R_{N-1}\circ ..\circ R_{1}(A),$ where $R_{N}$ can also be the element selector $\pi(q,\cdot)$. Each $R_{i}$ is associated with a different $k_i = |R_{i}(\cdot)|$, i.e., the number of elements the reranker returns.
Depending on the values of $k_i$, SR models with different trade-offs between accuracy and efficiency can be obtained.\footnote{The design of an end-to-end algorithm to learn the optimal parameter set for a given target trade-off is left as future work.}

\subsection{\astwo Definition}
The definition of \astwo directly follows from the definition of element selection of Section~\ref{AS}, where the query is a natural language question and the elements are answer sentence candidates retrieved with any approach, e.g., using a search engine.

%% file: sections/methodology.tex
\begin{figure}[t]
\centering
\includegraphics[width=0.9\columnwidth]{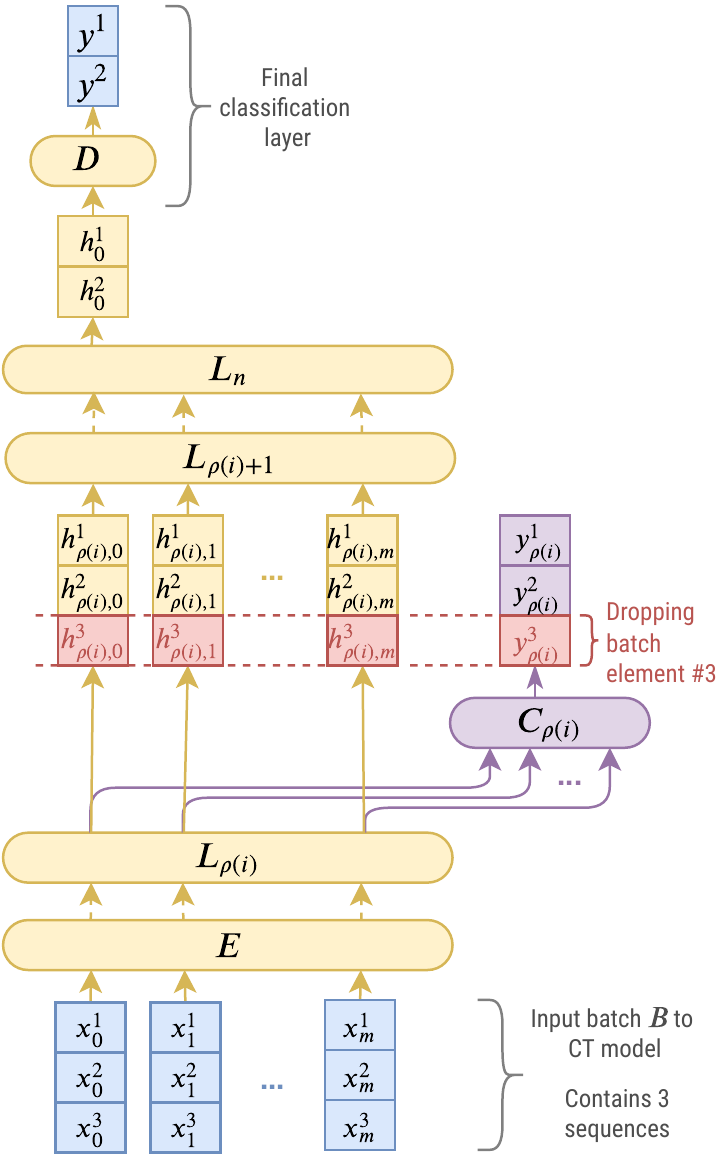}
\caption{
	A visual representation of the Cascade Transformer (CT) model proposed in this paper.
	Components in \textbf{\color{ylw}{yellow}} represent layers of a traditional transformer model, while elements in \textbf{\color{prp}{purple}} are unique to CT;
	input and outputs of the model are shown in \textbf{\color{ble}{blue}}.
	In this example, \textbf{\color{rdd}{drop rate $\bm{\alpha}\mathbf{=0.4}$}} causes sample $X^3$ to be removed by partial classifier $C_{\rho(i)}$.
}
\label{fig:cascade_regular}
\vspace{-.5em}
\end{figure}

\section{SR with transformers}
\label{sec:methodology}
In this section, we explain how to exploit the hierarchical architecture of a traditional transformer model to build an SR model.
First, we briefly recap how traditional transformer models (we refer to them as ``\textit{monolithic}'') are used for sequence classification, and how to derive a set of sequential rerankers from a pre-trained transformer model (Section~\ref{sub:monolithic_transformer_models}).
Then, we introduce our Cascade Transformer (CT) model, a SR model that efficiently uses partial encodings of its input to build a set of sequential rerankers $R_i$ (Section~\ref{sub:cascade_transformer_models}).
Finally, we explain how such model is trained and used for inference in sections~\ref{subsub:training}~and~\ref{subsub:inference}, respectively.

\subsection{Monolithic Transformer Models} %
\label{sub:monolithic_transformer_models}

We first briefly describe the use of transformer models for sequence classification. We call them \textit{monolithic} as, for all input samples, the computation flows from the first until the last of their layers.

Let ${\cal T}$ = $\{E; L_1, L_2, \ldots, L_n \}$ be a standard stacked transformer model \cite{vaswani2017attention}, where $E$ is the embedding layer, and $L_i$ are the transformer layers\footnote{That is, an entire transformer block, constituted by layers for multi-head attention, normalization, feed forward processing and positional embeddings.} generating contextualized representations for an input sequence;
$n$ is typically referred to as the depth of the encoder, i.e., the number of layers. Typical values for $n$ range from 12 to 24, although more recent works have experimented with up to 72 layers \cite{shoeybi2019megatronlm}.
${\cal T}$ can be pre-trained on large amounts of unlabeled text using a masked \cite{devlin2019bert,liu2019roberta} or autoregressive  \cite{yang2019xlnet,radford2019language} language modeling objective.

Pre-trained language models are fine-tuned for the target tasks using additional layers and data, e.g., a fully connected layer is typically stacked on top of ${\cal T}$ to obtain a sentence classifier.
Formally, given a sequence of input symbols\footnote{For ranking tasks, the sequence of input symbols is typically a concatenation of the query $q$ and a candidate $a_j$. In order for the model to distinguish between the two, a special token such as ``\texttt{[SEP]}'' or ``\texttt{</s>}'' is used. Some models also use a second embedding layer to represent which sequence each symbol comes from.}, $X = \{x_0, x_1, \ldots, x_m\}$, an encoding $H = \{h_0, h_1, \ldots, h_m\}$ is first obtained by recursively applying $H_i$ to the input:
\vspace{-.3em}
$$ H_0 = E(X),  H_i = L_i(H_{i-1})  \quad \forall i=1, \ldots, n,\vspace{-.5em}$$
where $H=H_n$.
Then, the first symbol of the input sequence\footnote{Before being processed by a transformer model, sequences are typically prefixed by a start symbol, such as ``\texttt{[CLS]}'' or ``\texttt{<s>}''.
This allows transformer models to accumulate knowledge about the entire sequence at this position without compromising token-specific representations \cite{devlin2019bert}.} is fed into a sequence of dense feed-forward layers $D$ to obtain a final output score, i.e., $y =D(h_0)$.
$D$ is fine-tuned together with the entire model on a task-specific dataset (a set of question and candidate answer pairs, in our case).

\subsection{Transformer-based Sequential Reranker (SR) Models} %
\label{sub:sr_transformer_models}

Monolithic transformers can be easily modified or combined to build a sequence of rerankers as described in Seciton~\ref{SR}.
In our case, we adapt an existing monolithic~${\cal T}$ to obtain a sequence of $N$~rerankers~$R_{i}$.
Each~$R_{i}$ consists of encoders from ~${\cal T}$ up to layer $\rho(i)$, followed by a classification layer $D_{i}$, i.e., $R_{i} = \{E; L_1, \ldots, L_{\rho(i)}, D_{i}\}$.
For a sequence of input symbols $X$, all rerankers in the sequence are designed to predict $p(q,a)$, which we indicate as $R_i(X) = y_{\rho(i)}$.
All rerankers in $SR_N$ are trained independently on the target data.

In our experiments, we obtained the best performance by setting $N=5$ and  using the following formula to determine the architecture of each reranker~$R_i$:\vspace{-.5em}
$$\rho(i)=4+2 \cdotp (i-1) \hspace{2em} \forall i=\{1,\ldots, 5\}\vspace{-.5em}$$
In other words, we assemble sequential reranker~SR$_5$ using five rerankers built with transformer models of 4, 6, 8, 10 and 12 layers, respectively.
This choice is due to the fact that our experimental results seem to indicate that the information in layers 1 to 3 is not structured enough to achieve satisfactory classification performance for our task.
This observation is in line with recent works on the effectiveness of partial encoders for semantic tasks similar to \astwo \cite{peters2019tune}.

\subsection{Cascade Transformer (CT) Models} %
\label{sub:cascade_transformer_models}

During inference, monolithic transformer models evaluate a sequence~$X$ through the entire computation graph to obtain the classification scores~$Y$.
This means that when using $SR_N$, examples are processed multiple times by similar layers for different $R_i$, e.g., for $i=1$, all $R_i$ compute the same operations of the first $\rho(i)$ transformer layers, for $i=2$, $N-1$ rerankers compute the same $\rho(i)-\rho(i+1)$, layers and so on.
A more computationally-efficient approach is to share all the common transformer blocks between the different rerankers in $SR_N$.

We speed up this computation by using one transformer encoder to implement all required $R_i$. This can be easily obtained by adding a classification layer $C_{\rho(i)}$ after each $\rho(i)$ layers (see Figure~\ref{fig:cascade_regular}).
Consequently, given a sample $X$, the classifiers $C_{\rho(i)}$ produces scores $y_{\rho(i)}$ only using a partial encoding.
To build a CT model, we use each $C_{\rho(i)}$ to build rerankers $R_i$, and select the top $k_i$ candidates to score with the subsequent rerankers  $R_{i+1}$.
We use the same setting choices of $N$ and $\rho(i)$ described in Section~\ref{sub:sr_transformer_models}.

Finally, we observed the best performance when all encodings in $H_{\rho(i)}$ are used as input to partial classifier~$C_{\rho(i)}$, rather than just the partial encoding of the classification token~$h_{\rho(i), 0}$.
Therefore, we use their average to obtain score $y_{\rho(i)} = C_{\rho(i)}(\frac{1}{m} \sum_{l=1,..,m} h_{\rho(i),l})$,
In line with \citet{kovaleva2019revealing}, we hypothesize that, at lower encoding layers, long dependencies might not be properly accounted in~$h_{\rho(i), 0}$.
However, in our experiments, we found no benefits in further parametrizing this operation, e.g., by either using more complex networks or weighting the average operation.

\subsubsection{Training CT}
\label{subsub:training}
The training of the proposed model is conducted in a multi-task fashion.
For every mini-batch, we randomly sample one of the rankers $R_i$ (including the final output ranker), calculate its loss against the target labels, and back-propagate its loss throughout the entire model down to the embedding layers.
We experimented with several more complex sampling strategies, including a round-robin selection process and a parametrized bias towards early rankers for the first few epochs, but we ultimately found that uniform sampling works best.
We also empirically determined that, for all classifiers $C_\rho(i)$, backpropagating the loss to the input embeddings, as opposed to stopping it at layer $\rho(i - 1)$, is crucial to ensure convergence.
A possible explanation could be: enabling each classifier to influence the input representation during backpropagation ensures that later rerankers are more robust against variance in partial encodings, induced by early classifiers.
We experimentally found that if the gradient does not flow throughout the different blocks, the development set performance for later classifiers drops when early classifiers start converging.

\subsubsection{Inference}
\label{subsub:inference}
Recall that we are interested in speeding up inference for classification tasks such as answer selection, where hundreds of candidates are associated with each question.
Therefore, we can assume without loss of generality that each batch of samples $B=\{X^1,\ldots, X^b\}$ contains candidate answers for the same question.
We use our partial classifiers to throw away a fraction $\alpha$ of candidates, to increase throughput.
That is, we discard $k_i = \lfloor \alpha \cdotp k_{i-1} \rfloor$ candidates, where $\lfloor \cdot \rfloor$ rounds $\alpha \cdotp k_{i-1}$ down to the closest integer.

For instance, let $\alpha = 0.3$, batch size $b=128$; further, recall that, in our experiments, a CT consists of 5 cascade rerankers.
Then, after layer 4, the size of the batch gets reduced to~$90$ ($\lfloor 0.3 \cdotp 128 \rfloor = 38$ candidates are discarded by the first classifier).
After the second classifier (layer 6), $\lfloor 0.3 \cdotp 90 \rfloor = 27$ examples are further removed, for an effective batch size of~$63$.
By layer 12, only~$31$ samples are left, i.e., the instance number scored by the final classifier is reduced by more than 4 times.

Our approach has the effect of improving the throughput of a transformer model by reducing the average batch size during inference:
the throughput of any neural model is capped by the maximum number of examples it can process in parallel (i.e., the size of each batch), and said number is usually ceiled by the amount of memory available to the model (e.g., RAM on GPU).
The monolithic models have a constant batch size at inference; however, because the batch size for a cascade model varies while processing a batch, we can size our network with respect to its average batch size, thus increasing the number of samples we initially have in a batch.
In the example above, suppose that the hardware requirement dictates a maximum batch size of~$84$ for the monolithic model. As the average batch size for the cascading model is $(4 \cdotp 128 + 2\cdotp 90 + 2\cdotp 63 + 2\cdotp 44 + 2\cdotp 28) /12 = 80.2 < 84$, we can process a batch of 128 instances without violating memory constrains, increasing throughput by~$52\%$.

We remark that using a fixed $\alpha$ is crucial to obtain the performance gains we described:
if we were to employ a score-based thresholding approach (that is, discard all candidates with score below a given threshold), we could not determine the size of batches throughout the cascade, thus making it impossible to efficiently scale our system.
On the other hand, we note that nothing in our implementations prevents potentially correct candidates from being dropped when using CT.
However, as we will show in Section~\ref{exp}, an opportune choice of a threshold and good accuracy of early classifiers ensure high probability of having at least one positive example in the candidate set for the last classifier of the cascade.

%% file: sections/experimental-setup.tex
\section{Experiments}
\label{exp}

We present three sets of experiments designed to evaluate CT.
In the first (Section~\ref{sub:stability_of_cascade_training}), we show that our proposed approach without any selection produces comparable or superior results with respect to the state of the art of \astwo, thanks to its stability properties; in the second (Section~\ref{sub:effectiveness_of_cascading}), we compare our Cascade Transformer with a vanilla transformer, as well as a sequence of transformer models trained independently; finally, in the third (Section~\ref{sub:drop_ratio_tuning}), we explore the tuning of the drop ratio, $\alpha$.

\begin{table}[t]
\centering
\small
\begin{tabular}[t]{llrrrr}
\toprule
&
& {\scriptsize \textbf{ASNQ}}
& {\scriptsize \textbf{GPD}}
& {\scriptsize \textbf{TRECQA}}
& {\scriptsize \textbf{WikiQA}}	\\
\midrule
\multirow{3}{*}{\STAB{\rotatebox[origin=c]{90}{\textsc{train}}}}
& Questions			& 57,242 		& 1,000			& 1,227				& 873				\\
& Avg cand.			& 413.3  		& 99.8			& 39.2				& 9.9				\\
& Avg corr.  		& 1.2 	 		& 4.4			& 4.8				& 1.2				\\
\midrule
\multirow{3}{*}{\STAB{\rotatebox[origin=c]{90}{\textsc{dev}}}}
& Questions			& 1,336			& 340			& 65				& 126				\\
& Avg cand.			& 403.6			& 99.7			& 15.9				& 9.0				\\
& Avg corr.  		& 3.2			& 2.85			& 2.9				& 1.1				\\
\midrule
\multirow{3}{*}{\STAB{\rotatebox[origin=c]{90}{\textsc{test}}}}
& Questions			& 1,336			& 440			& 68				& 243				\\
& Avg cand.			& 400.5			& 101.1			& 20.0				& 9.7				\\
& Avg corr.  		& 3.2			& 8.13			& 3.4				& 1.2				\\
\bottomrule
\end{tabular}
\caption{Datasets statistics: ASNQ and GPD have more sentence candidates than TRECQA and WikiQA.}
\label{tab:dataset_stats}
\vspace{-1em}
\end{table}

\subsection{Datasets} %
\label{sec:datasets}

\paragraph{TRECQA \& WikiQA}
Traditional benchmarks used for \astwo, such as TRECQA \cite{wang-etal-2007-jeopardy} and WikiQA \cite{yang-etal-2015-wikiqa}, typically contain a limited number of candidates for each question.
Therefore, while they are very useful to compare accuracy of \astwo systems with the state of the art, they do not enable testing large scale passage reranking, i.e., inference on hundreds or thousand of answer candidates.  Therefore, we evaluated our approach (Sec.~\ref{sub:cascade_transformer_models}) on two datasets: ASNQ, which is publicly available, and our GPD dataset.
We still leverage TRECQA and WikiQA to show that that our cascade system has comparable performance to state-of-the-art transformer models when no filtering is applied. %

\paragraph{ASNQ} The Answer Sentence Natural Questions dataset \cite{garg2019tanda} is a large collection (23M samples) of question-answer pairs, which is two orders of magnitude larger than most public \astwo datasets.
It was obtained by extracting sentence candidates from the Google Natural Question (NQ) benchmark~\cite{47761}.
Samples in NQ consists of tuples $\langle\texttt{question},$ $\texttt{answer}_\texttt{long},$ $\texttt{answer}_\texttt{short},$ $\texttt{label}\rangle$, where $\texttt{answer}_\texttt{long}$ contains multiple sentences, $\texttt{answer}_\texttt{short}$ is fragment of a sentence, and \texttt{label} is a binary value indicating whether $\texttt{answer}_\texttt{long}$ is correct.
The positive samples were obtained by extracting sentences from $\texttt{answer}_\texttt{long}$ that contain $\texttt{answer}_\texttt{short}$;
all other sentences are labeled as negative.
The original release of ANSQ\footnote{\url{https://github.com/alexa/wqa_tanda}} only contains train and development splits; we further split the dev.~set to both have dev.~and test sets.

\paragraph{GPD} The General Purpose Dataset is  part of our efforts to study large scale web QA and evaluate performance of \astwo systems.
We built GPD using a search engine to retrieve up to 100 candidate documents for a set of given questions. Then, we extracted all candidate sentences from such documents, and rank them using a vanilla transformer model, such as the one described in Sec.~\ref{sub:monolithic_transformer_models}.
Finally, the top 100 ranked sentences were manually annotated as correct or incorrect answers.

We measure the accuracy of our approach on ASNQ and GPD using four metrics: Mean Average Precision (MAP), Mean Reciprocal Rank (MRR), Precision at 1 of ranked candidates (P@1), and Normalized Discounted Cumulative Gain at 10 of retrieved candidates (nDCG@10).
While the first two metrics capture the overall system performance, the latter two are better suited to evaluate systems with many candidates, as they focus more on Precision.
For WikiQA and TRECQA, we use MAP and MRR.

\subsection{Models and Training} %
\label{sub:models}

Our models are fine-tuned starting from a pre-trained RoBERTa encoder \cite{liu2019roberta}.
We chose this transformer model over others due to its strong performance on answer selection tasks \cite{garg2019tanda}.
Specifically, we use the \textsc{Base} variant (768-dimensional embeddings, 12 layers, 12 heads, and 3072 hidden units), as it is more appropriate for efficient classification.
When applicable\footnote{When fine-tuning on GPD, TRECQA, and WikiQA, we perform a ``transfer'' step on ASNQ before ``adapting'' to our target dataset; for ASNQ, we directly fine-tune on it.}, we fine-tune our models using the two-step ``transfer and adapt'' (\tanda) technique introduced by \citet{garg2019tanda}.

As mentioned in Section~\ref{sub:cascade_transformer_models}, we optimize our model in a multi-task setting; that is, for each mini-batch, we randomly sample one of the output layers of the CT classifiers to backpropagate its loss to all layers below.

While we evaluated different sampling techniques, we found that a simple uniform distribution is sufficient and allows the model to converge quickly.

Our models are optimized using Adam \cite{kingma2014adam} using triangular learning rate \cite{smith2017cyclical} with a $4,000$ updates ramp-up\footnote{On ASNQ, it is roughly equivalent to $\tilde{950k}$ samples or about $4\%$ of the training set.}, and a peak learning rate $l_r=1e^{-6}$.
Batch size was set to up to $2,000$ tokens per mini-batch for CT models.
For the partial and final classifiers, we use 3-layers feed-forward modules with with 768 hidden units and $tanh$ activation function.
Like the original BERT implementation, we use dropout value of 0.1 on all dense and attention layers.
We implemented our system using MxNet 1.5 \cite{chen2015mxnet} and GluonNLP 0.8.1 \cite{guo2019gluoncv} on a machine with 8 NVIDIA Tesla V100 GPUs, each with 16GB of memory.

%% file: sections/results.tex
\begin{table}[t]
\center
\small
\begin{tabular}{l|ll|ll}
\toprule
\multirow{ 2}{*}{\textbf{Model}}& \multicolumn{2}{c|}{\textbf{WikiQA}} 	& \multicolumn{2}{c}{\textbf{TRECQA}}	\\
& {\textsc{map}}				& {\textsc{mrr}}				& {\textsc{map}}			& {\textsc{mrr}}	\\
\midrule
CA1 {\tiny \cite{wang2016compare}}	& 74.3 			& 75.4 				& --			& -- 						\\
CA2 {\tiny \cite{DBLP:journals/corr/abs-1905-12897}}& 83.4 			& 84.8 				& 87.5 			& 94.0 						\\
$\tanda_\textsc{base}$ {\tiny \cite{garg2019tanda}}	& 88.9 			& 90.1 				& 91.4 			& 95.2 						\\
\midrule
4 layers \tanda				& 80.5	& 80.9 	& 77.2 	& 83.1 \\
6 layers \tanda				& 82.1	& 82.9	& 78.5	& 88.4 \\
8 layers \tanda				& 85.7	& 86.7	& 88.2	& 94.7 \\
10 layers \tanda			& 89.0	& 90.0	& 90.5	& 95.9 \\
Our $\tanda_\textsc{base}$	& 89.1	& 90.1	& 91.6	& 96.0	\\
\midrule
CT ($4$ layers, $\alpha=0.0$)	& 60.1				& 60.2 			& 67.9		 	& 74.7						\\
CT ($6$ layers, $\alpha=0.0$)	& 79.8				& 80.3 			& 89.7		 	& 95.0						\\
CT ($8$ layers, $\alpha=0.0$)	& 84.8				& 85.4 			& 92.3		 	& 95.3						\\
CT ($10$ layers, $\alpha=0.0$)	& 89.7				& 89.8 			& 92.3		 	& 95.6						\\
CT ($12$ layers, $\alpha=0.0$)	& \textbf{89.9}		& \textbf{91.0}	& \textbf{92.4}	& \textbf{96.7}				\\
\bottomrule
\end{tabular}
\caption{
Comparison on two \astwo academic datasets.
With the exception of a 4-layer transformer, both the partial and final classifiers from CT achieve comparable or better performance than state of the art models.
}
\label{tab:wiki_trec}
\vspace{-.5em}
\end{table}

\begin{table*}[t]
\ra{1.1}
\small
\centering
\begin{tabular}{llc|cccc|cccc|r}
\toprule
\multirow{2}{*}{\textbf{Method}} & \multirow{2}{*}{\textbf{Model}} & \multirow{2}{*}{$\bm{\alpha}$}  &  \multicolumn{4}{c|}{\textbf{ASNQ}} &  \multicolumn{4}{c|}{\textbf{GPD}}  & \multirow{2}{*}{\makecell[c]{\scriptsize \textbf{Cost reduction} \\ \scriptsize \textbf{per batch}}} \\
&&& {\scriptsize MAP} & {\scriptsize nDCG@10} & {\scriptsize P@1} & {\scriptsize MRR} & {\scriptsize MAP} & {\scriptsize nDCG@10} & {\scriptsize P@1} & {\scriptsize MRR} & \\
\midrule
\multirow{5}{*}{\makecell[l]{Monolithic \\ transformer \\ (MT)}}
& 4 layers {\scriptsize \tanda}	& -- 	& 31.5 	& 30.8 	& 25.9 	& 30.8	& 38.9 	& 50.1 & 40.8 & 54.0 & $-67$\% \\
& 6 layers {\scriptsize \tanda}	& -- 	& 60.2 	& 58.7 	& 47.2 	& 59.2	& 51.4 	& 64.1 & 56.1 & 67.6 & $-50$\% \\
& 8 layers {\scriptsize \tanda}	& -- 	& 63.9 	& 62.2 	& 49.2 	& 62.4	& 56.3 	& 68.7 & 61.2 & 70.4 & $-33$\% \\
& 10 layers {\scriptsize \tanda} 	& -- 	& 65.3 	& 64.5 	& 52.0 	& 64.1	& 57.2	& 71.3 & 64.9 & 72.7 & $-20$\% \\
\cdashline{2-12}
& $\tanda_\textsc{base}$
					& -- 	& 65.5 	& 65.1 	& 52.1 	& 64.7	& {\bf58.0} & {\bf72.2} & {\bf67.5} & 76.8	& {\textit{baseline}} \\
\midrule
\multirow{3}{*}{\makecell[l]{Sequential\\ Ranker (SR)}}
& \multirow{3}{*}{\makecell[l]{\small MT models, \\ \small 4 to 12 layers, \\ \small in sequence}}
				& 0.3	& 65.4	& 65.1	& 52.1	& 64.8	& 55.8	& 70.2 & 66.2 & 74.3 & $+53$\% \\
& 				& 0.4	& 64.9	& 64.2	& 51.6	& 64.2	& 53.8	& 69.6 & 65.6 & 73.0 & $+18$\% \\
& 				& 0.5	& 64.6	& 63.4	& 50.8	& 63.5	& 52.2	& 68.4 & 63.0 & 72.3 & $-10$\% \\
\midrule
\multirow{8}{*}{\makecell[l]{Cascade\\ transformer \\ (CT)}}
& 4 layers CT		& 0.0 	& 22.0	& 19.3	& 10.2	& 18.3	& 32.7	& 38.9	& 35.2 & 42.6 & $-67$\% \\
& 6 layers CT		& 0.0 	& 49.1	& 47.2	& 32.7	& 47.7	& 44.8	& 56.0	& 47.3 & 58.5 & $-50$\% \\
& 8 layers CT		& 0.0 	& 62.8	& 61.5	& 48.7	& 61.9	& 53.8	& 71.7	& 61.2 & 69.1 & $-33$\% \\
& 10 layers CT		& 0.0	& 65.6	& 65.1	& 53.0	& 65.2	& 55.8	& 72.0	& 63.1 & 72.1 & $-20$\% \\
\cdashline{2-12}
& \multirow{4}{*}{\makecell[l]{Full CT \\ (12 layers)}}
				& 0.0	& {\bf66.3}	& {\bf66.1}	& {\bf53.2}	& {\bf65.4}	& 57.8 & 71.9 & {\bf67.5} & {\bf76.9} & {$-0$\%} \\
&				& 0.3	& 65.3	& 65.3	& 52.9	& 65.3	& 55.7	& 69.8	& 66.2 & 75.1 & $-37$\% \\
&				& 0.4	& 64.8	& 65.0	& 52.5	& 64.8	& 52.8	& 68.6	& 65.6 & 74.3 & $-45$\% \\
&				& 0.5	& 64.1	& 65.0	& 52.4	& 64.5	& 50.2	& 66.1	& 62.4 & 72.9 & $-51$\% \\
 \bottomrule
\end{tabular}
\caption{Comparison of Cascade Transformers with other models on the ASNQ and GPD datasets.
``Monolithic transformer'' refers to a single transformer model trained independently; ``sequential ranker'' (ST) is a sequence of monolithic transformer models of size $4, 6, \ldots, 12$ trained independently; and ``Cascade Transformer'' (CT) is the approach we propose.
This can train models that equal or outperform the state of the art when no drop is applied (i.e., $\alpha = 0.0$); with drop, they obtain the same performance with $37\%$ to $51\%$ fewer operations.}
\label{tab:gpd_asnq}
\end{table*}

\subsection{Stability Results of Cascade Training} %
\label{sub:stability_of_cascade_training}

In oder to better assess how our training strategy for CT models compare with a monolithic transformer, we evaluated the performance of our system on two well known \astwo datasets, WikiQA and TRECQA.
The results of these experiments are presented in Table~\ref{tab:wiki_trec}.
Note how, in this case, we are not applying any drop to our cascade classifier, as it is not necessary on this dataset: all sentences fit comfortably in one mini batch (see dataset statistics in Table~\ref{tab:dataset_stats}), so we would not observe any advantage in pruning candidates.
Instead, we focus on evaluating how our training strategy affects performance of partial and final classifiers of a CT model.

Our experiment shows that classifiers in a CT model achieve competitive performance with respect to the state of the art: our 12-layer transformer model trained in cascade outperforms $\tanda_\textsc{base}$ by $0.8$ and $0.9$ absolute points in MAP ($0.9$ and $0.7$ in MRR).
10, 8, and 6 layer models are equally comparable, differing at most by $2.3$ absolute MAP points on WikiQA, and outscoring \tanda by up to $11.2$ absolute MAP points on TRECQA.
However, we observed meaningful differences between the performance of the 4-layers cascade model and its monolithic counterparts.
We hypothesize that this is due to the fact that lower layers are not typically well suited for classification when used as part of a larger model \cite{peters2019tune};
this observation is reinforced by the fact that the 4 layers \tanda model shown in Table~\ref{tab:wiki_trec} takes four times the number of the iterations of any other model to converge to a local optimum.

Overall, these experiments show that our training strategy is not only effective for CT models, but can also produce smaller transformer models with good accuracy without separate fine-tuning.

\subsection{Results on Effectiveness of Cascading} %
\label{sub:effectiveness_of_cascading}

The main results for our CT approach are presented in Table~\ref{tab:gpd_asnq}:
we compared it with (\textit{i}) a state-of-the-art monolithic transformer ($\tanda_\textsc{base}$), (\textit{ii}) smaller, monolithic transformer models with 4-10 layers, and (\textit{iii}) a sequential ranker (SR) consisting of 5 monolithic transformer models with $4, 6, 8, 10$ and $12$ layers trained independently.
For CT, we report performance of each classifier individually (layers 4 up to 12, which is equivalent to a full transformer model).
We test SR and CT with drop ratio 30\%, 40\%, 50\%.
Finally, for each model, we report the relative cost per batch compared to a \textit{base} transformer model with 12 layers.

Overall, we observed that our cascade models are competitive with monolithic transformers on both ASNQ and GPD datasets.
In particular, when no selection is applied ($\alpha=0.0$), a 12 layer cascade model performs equal or better to $\tanda_\textsc{Base}$:
on ASNQ, we improve P@1 by 2.1\% ($53.2$ vs $52.1$), and MAP by 1.2\% ($66.3$ vs $65.5$);
on GDP, we achieve the same P@1 ($67.5$), and a slightly lower MAP ($57.8$ vs $58.0$).
This indicates that, despite the multitasking setup, out method is competitive with the state of the art.

A drop rate $\alpha > 0.0$ produces a small degradation in accuracy, at most, while significantly reducing
the number of operations per batch ($-37\%$).
In particular, when $\alpha = 0.3$, we achieve less than 2\% drop in P@1 on GPD, when compared to $\tanda_\textsc{Base}$; on ANSQ, we slightly improve over it ($52.9$ vs $52.1$). We observe a more pronounced drop in performance for MAP, this is to be expected, as intermediate classification layers are designed to drop a significant number of candidates.

For larger values of $\alpha$, such as $0.5$, we note that we achieve significantly better performance than monolithic transformer of similar computational cost.
For example, CT achieves an~$11.2$\% improvement in P@1 over a 6-layers \tanda model ($62.4$ vs $56.1$) on GPD; a similar improvement is obtained on ANSQ ($+11.0\%$, $52.4$ vs $47.2$).

Finally, our model is also competitive with respect to a sequential transformer with equivalent drop rates, while being between 1.9 to 2.4 times more efficient.
This is because an SR model made of independent \tanda models cannot re-use encodings generated by smaller models as  CT does.

\subsection{Results on Tuning of Drop Ratio $\bm{\alpha}$} %
\label{sub:drop_ratio_tuning}

\begin{figure}[t]
\centering
\includegraphics[width=1.0\columnwidth]{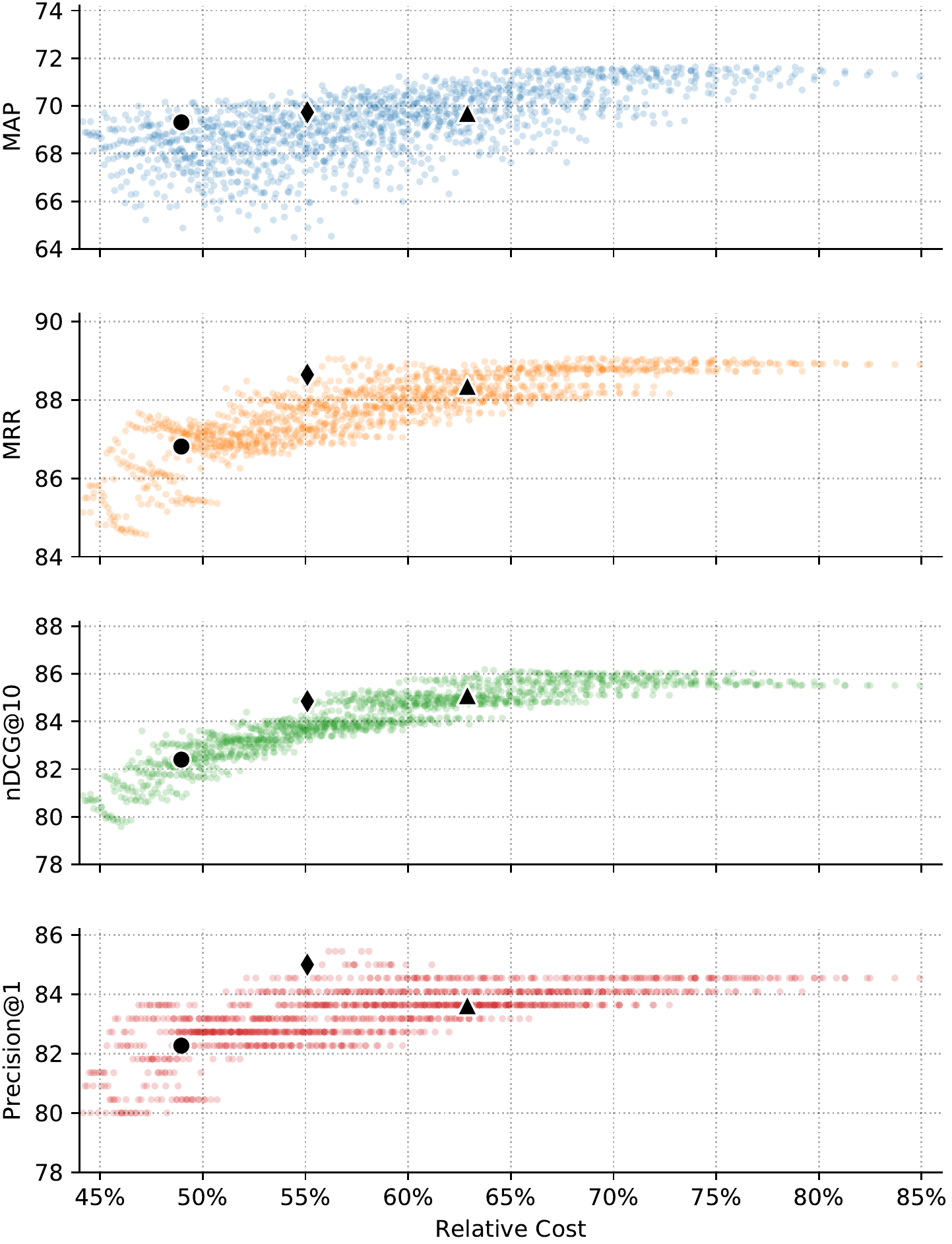}
\caption{
Grid search plot on the GPD validation set.
Each point corresponds to a configuration of drop ratios $\{\alpha_{p_1}, \ldots, \alpha_{p_4}\}$ with $\alpha_{p_k} \in \{0.1, 0.2, \ldots, 0.6\}$; values on the $x$-axis represent the relative computational cost per batch of a configuration compared to $\tanda_\textsc{base}$.
The three runs reported in Table~\ref{tab:gpd_asnq} correspond to $\blacktriangle$ ($\alpha=0.3$), $\blacklozenge$ ($\alpha=0.4$), and $\CIRCLE$ ($\alpha=0.5$).}
\label{fig:gpd_tuning}
\end{figure}

Finally, we examined how different values for drop ratio $\alpha$ affect the performance of CT models.
In particular, we performed an exhaustive grid-search on a CT model trained on the GPD dataset for drop ratio values $\{\alpha_{p_1}, \alpha_{p_2}, \alpha_{p_3}, \alpha_{p_4}\}$, with $\alpha_{p_k} \in \{0.1, 0.2, \ldots, 0.6\}$.
The performance is reported in Figure~\ref{fig:gpd_tuning} with respect to the relative computational cost per batch of a configuration when compared with a $\tanda_\textsc{base}$ model.

Overall, we found that CT models are robust with respect to the choice of $\left\{\alpha_{p_k}\right\}_{k=1}^{4}$.
We observe moderate degradation for higher drop ratio values (e.g., P@1 varies from $85.6$ to $80.0$).
Further, as expected, performance increases for models with higher computational cost per batch, although they taper off for CT models with relative cost $\geq 70\%$.
On the other hand, the grid search results do not seem to suggest an effective strategy to pick optimal values for $\left\{\alpha_{p_k}\right\}_{k=1}^{4}$, and, in our experiments, we ended up choosing the same values for all drop rates. In the future, we would be like to learn such values while training the cascade model itself.

%% file: sections/conclusions.tex
\section{Conclusions and Future Work}
\label{sec:conclusions}

This work introduces CT, a variant of the traditional transformer model designed to improve inference throughput.
Compared to a traditional monolithic stacked transformer model, our approach leverages classifiers placed at different encoding stages to prune candidates in a batch and improve model throughput.
Our experiments show that a CT model not only achieves comparable performance to a traditional transformer model while reducing computational cost per batch by over $37\%$, but also that our training strategy is stable and jointly produces smaller transformer models that are suitable for classification when higher throughput and lower latency goals must be met.
In future work, we plan to explore techniques to automatically learn where to place intermediate classifiers, and what drop ratio to use for each one of them.